# Playing Tic-Tac-Toe Games with Intelligent Single-pixel Imaging

Shuming Jiao[1,*], Jiaxiang Li[2], Wei Huang[3], Zibang Zhang[2,4]

1.Peng Cheng Laboratory, Shenzhen 518055, Guangdong, China

2.Department of Optoelectronic Engineering, Jinan University, Guangzhou 510632, Guangdong, China

3.College of Electrical and Information Engineering, Hunan University, Changsha, Hunan, China

4. tzzb@jnu.edu.cn

Corresponding author*: jiaoshm@pcl.ac.cn

*Abstract:*

*Single-pixel imaging (SPI) is a novel optical imaging technique by replacing a two-dimensional pixelated sensor with a single-pixel detector and pattern illuminations. SPI have been extensively used for various tasks related to image acquisition and processing. In this work, a novel non-image-based task of playing Tic-Tac-Toe games interactively is merged into the framework of SPI. An optoelectronic artificial intelligent (AI) player with minimal digital computation can detect the game states, generate optimal moves and display output results mainly by pattern illumination and single-pixel detection. Simulated and experimental results demonstrate the feasibility of proposed scheme and its unbeatable performance against human players.*

Single-pixel imaging (SPI) [1-3] is a novel optical imaging technique based on sequential pattern illuminations and single-pixel detection. As a comparison, a two-dimensional object image is captured by a conventional camera at one shot with a pixelated sensor array. In SPI, the object scene is illuminated by different projected structured light patterns in sequence, and the total light intensity is collected by a single-pixel detector at each time. Mathematically, the recorded light intensity is proportional to the inner product between an encode illumination pattern and the object image. Finally, a single-pixel intensity data sequence will be recorded and the object image can be computationally reconstructed from the sequence plus the illumination pattern data. SPI is a physical implementation of compressive sensing in an optical manner and the illumination patterns constitute the equivalent measurement matrix. A typical optical setup of a SPI system is shown in Fig. 1.

SPI has potential advantages of low sensor cost in some invisible wavebands and it is suitable for imaging tasks under some unconventional conditions. SPI has been applied for Lidar imaging [4], 3D imaging [5], scattering imaging [6], underwater imaging [7], microscopy [8], holography [9], night vision [10] and non-line-of-sight imaging [11]. SPI can also be leveraged for image processing tasks such as image classification [12-15], edge detection [16] and object tracking [17,18] since it can directly acquire compressed and transformed object image data. In recent years, SPI has been combined with emerging deep learning methods [19-24] extensively but they are still mainly focused on imaging and vision. As an inherent optical imaging system, the application of SPI is mostly limited to image acquisition and processing in most of the previous works stated above. It is favorable that other non-image-based information processing tasks are attempted and investigated as well based on the unique features of SPI. In this paper, SPI is used as an optical artificial intelligent (AI) machine for playing Tic-Tac-Toe games.

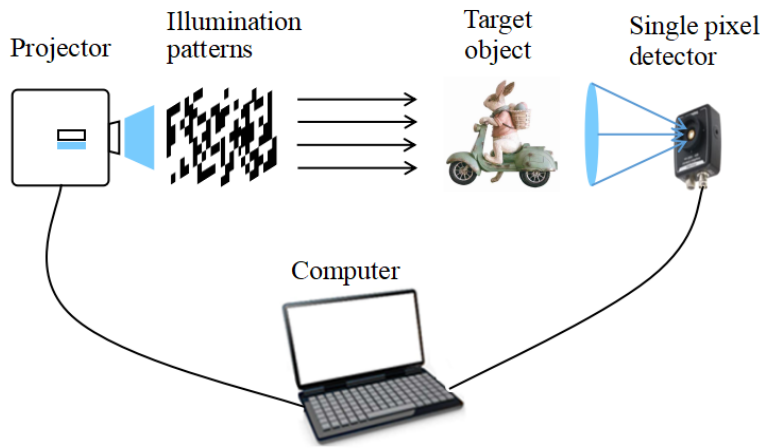

Fig. 1. Optical setup of a SPI system

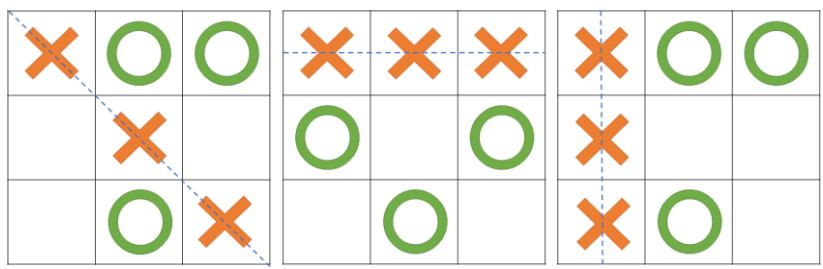

Fig. 2 Tic-Tac-Toe game (winning by connecting three symbols in a diagonal, horizontal or vertical line)

It is considered as a significant advance for building a man-made machine with certain intelligence to play interactive games with human players. For example, the world chess champion was first defeated by Deep Blue in 1997 [25] and the world Go champion was first defeated by AlphaGo in 2016 [26]. Compared with sophisticated chess and Go games, a simple board game like Tic-Tac-Toe [27] serves as a fundamental testing platform for preliminarily developed machine learning systems. In as early as 1952, Tic-Tac-Toe was played by the electronic EDSAC computer at the University of Cambridge [28]. In 1975, a perfect play of Tic-Tac-Toe was used to demonstrate the computational power of new Tinkertoy computer in Massachusetts Institute of Technology (MIT) [29]. The capability of playing Tic-Tac-Toe games was also employed to verify the feasibility of a new type of "chemical molecular computer" without electronics [28,30].

Tic-Tac-Toe is a game played on a three-by-three grid by two players, shown in Fig. 2. On the game board, all the grids have empty space initially. Then the two players alternately place a symbol (e.g. "X" denotes Player 1 and "O" denotes Player 2) of his/her own in an open square that has not been occupied. One player who has three symbols in a row, column or diagonal line first will be the winner. The game may possibly end with a draw.

Due to the simple rules and small-scale search space, all the possible combinations in a Tic-Tac-Toe game can be easily exhausted by a conventional computer in a short time. So for each possible game state, an absolutely optimal strategy for one player can be found with efficient algorithms such as Minimax algorithm [31]. The basic idea is that all the possible game states are represented as nodes and connected as a game tree. Each path in the tree is searched and a worst-case score is obtained against all possible opponent strategies. The Minimax algorithm will determine the optimal action (next move) by finding the root minimax value (best worst-case score) from leaf values. The algorithm can theoretically ensure that the computer player will never lose the game and the worst situation is a draw.

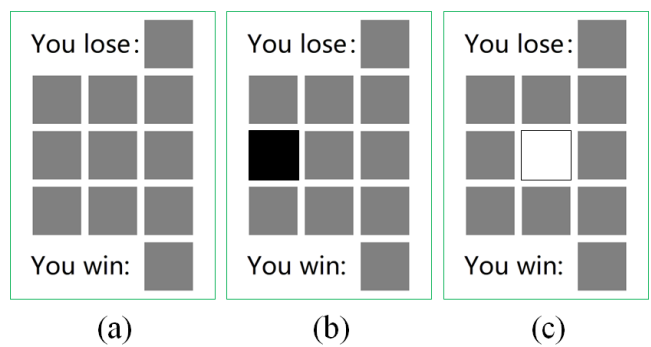

Fig. 3 (a) Printed Tic-Tac-Toe game board with a 3×3 gray square grid in this work; (b)human player's placing a symbol by attaching a black paper card over a gray square;(c)SPI player's placing a symbol by attaching a white paper card over a gray square;

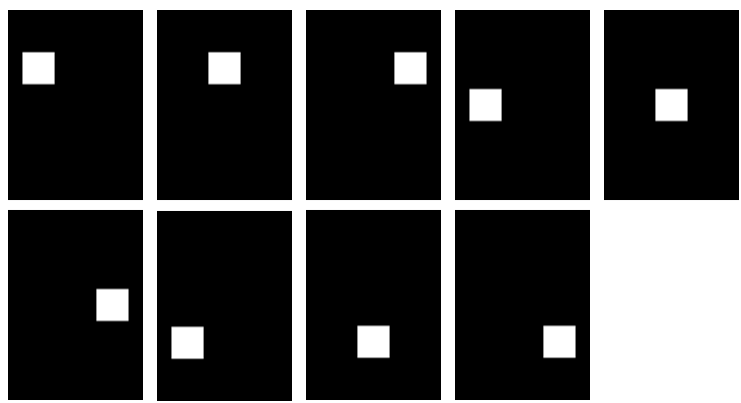

Fig. 4 Nine illumination patterns with an identical size as the board for detecting the game state.

A SPI system can be tailor-made as a Tic-Tac-Toe player in the following way. First, the system is required to detect the game state at any time with nine pattern illuminations. A printed game board with nine gray color squares in a grid is shown in Fig. 3. The symbols for human opponent player and the SPI player are black and white square paper cards respectively. If one player occupies an empty square with his/her symbol, the original gray color will be overlapped by black or white color by attaching the paper card on the board. At the detection stage, the projector will project nine sequential illumination patterns, shown in Fig. 4. In each pattern, only the area corresponding to one of the nine squares on the board is illuminated. Nine corresponding single-pixel intensity values can be recoded by the single-pixel detector. The intensity value can indicate whether each square has black color (maximum light absorption and minimum light reflection), gray color (medium light absorption and reflection) or white color (minimum light absorption and maximum light reflection) color, shown in Fig. 5. Consequently, the game state, represented by a vector containing nine elements, can be obtained. The vector element will have a value of 1 (occupied by human player), 2 (empty) or 3 (occupied by SPI player) corresponding to each square in the grid.

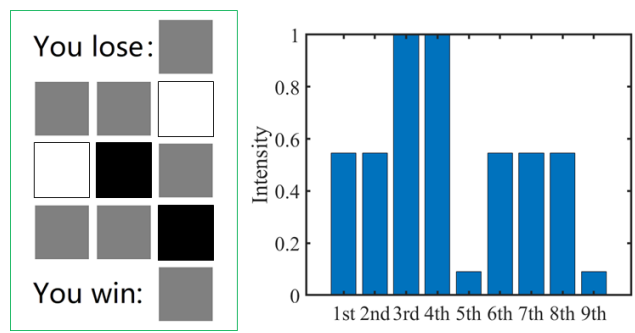

Fig. 5 Recorded single-pixel intensity sequence for a game state (lowest intensity for black squares, medium intensity for gray squares and maximum intensity for white squares)

Second, the system can generate the optimal action for the next move based on the current game state with minimal digital computation. Theoretically, the vector has totally $3^9$=19683 possibilities. In practice, the difference between the number of white symbols and black symbols is 0 or $\pm 1$ since two players take turns to place one symbol at each time. So only some of the vector value combinations among the 19683 ones are legal under this constraint, which may occur in real games. For all the legal game states, the optimal action (position of square for placing the symbol in) in the next move can be optimized in advanced off-line by the Minimax algorithm stated above. In addition, it will be evaluated whether the human player wins in the current game state and whether the SPI player will win after the current selected move. Then the pre-calculated results for all possible game states are stored in a lookup table. When the game is played on-line, the only digital processing step is result retrieval from the lookup table based on the detected game state.

Third, the display of SPI player's action and the human player's winning or losing status is implemented by pattern illumination as well. Totally 19 possible patterns will be used for the respective situation, shown in Fig. 6. The square where the SPI player intends to place a white symbol in the next move will be illuminated. The top and bottom squares on the board indicating the human player's winning or losing status will be illuminated as well if necessary.

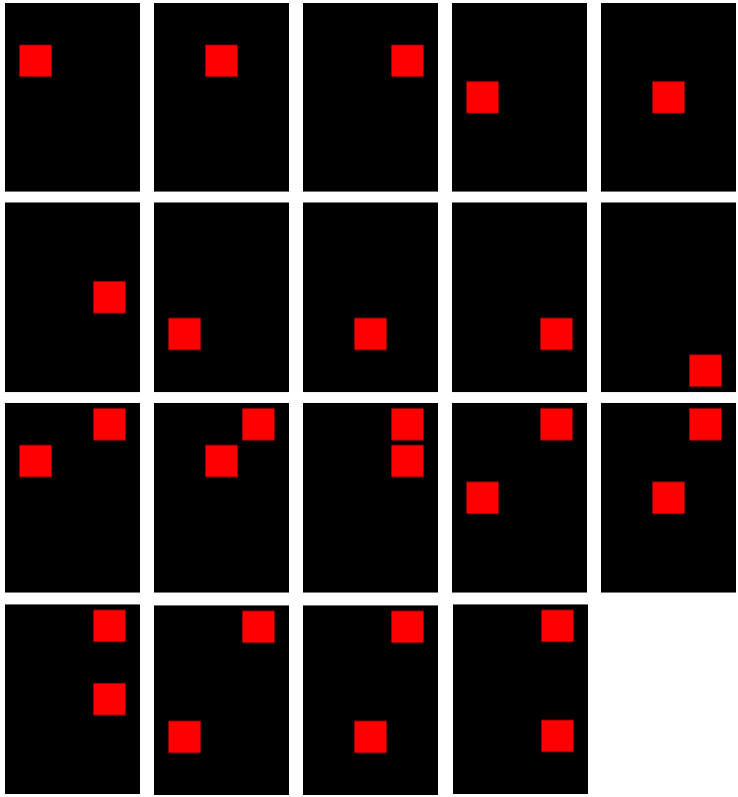

Fig. 6 Nineteen illumination patterns for displaying the output results (from left to right and from top to bottom: first nine patterns indicate the SPI player's next action; the tenth pattern indicates the human player already wins and there is no need for further action; the eleventh to nineteenth patterns indicate the SPI player's next action and the SPI player will be the winner after this action).

Simulation and optical experiment are conducted to verify our proposed scheme. In the simulation, 2000 virtual Tic-Tac-Toe games are played between our proposed SPI player against a computer random player. In 1000 games, the SPI player moves first and in the other 1000 games, the random player moves first. Each time the computer random player adopts the most straightforward strategy and randomly chooses an empty square to place the symbol in. The results in Table 1 show that our designed SPI player with pre-stored optimal strategies has a superior unbeatable performance compared with a computer random player. If the SPI player moves first, the wining probability will be even higher. It can win the game in most time, have a draw sometimes and never encounters a losing situation. Then ten volunteers of adult human players are invited to test our system. Each human player has two games with the SPI player. The SPI player moves first in one game and the human player moves first in the other game. The results in Table 2 show that there is still no losing situation against real human players. The SPI player can defeat human player in some games even though many games end with a draw. The overall intelligence of this system for playing Tic-Tac-Toe games is at least comparable to or better than average real human players

Table 1 Simulated Tic-Tac-Toe game results between our proposed SPI player with a computer random player

| SPI Player | Win | Draw | Lose |
|---|---|---|---|
| Move First | 97.6% | 2.4% | 0% |
| Move Second | 84.3% | 15.7% | 0% |

Table 2 Simulated Tic-Tac-Toe game results between our proposed SPI player with real human players

| SPI Player | Win | Draw | Lose |
|---|---|---|---|
| Move First | 5 | 5 | 0 |
| Move Second | 1 | 9 | 0 |

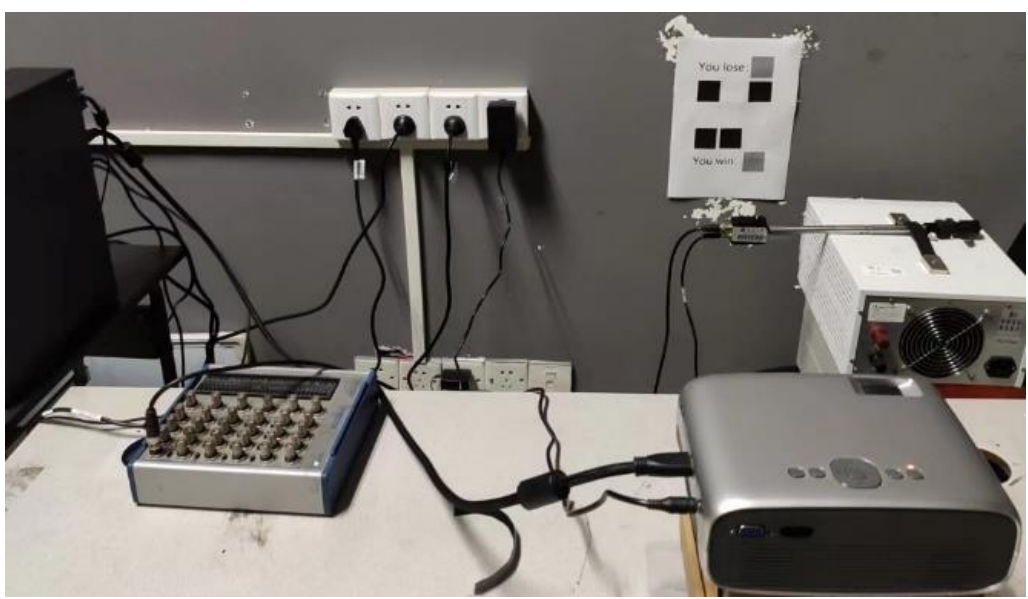

Fig. 7 Experimental SPI setup for this work

In the optical experiment, an optical experimental setup is shown in Fig. 7. A Philips NeoPix Easy 2+ NPX442 commercial projector is used for pattern illumination. A Thorlabs 100A-EC PDA (photo-diode array) is used as the single-pixel detector and its recorded data are collected by a NI-USB-6343 data acquisition card connected to a computer. Two Tic-Tac-Toe games are played between the SPI system and a human player as demonstration examples. The processes of two games are shown in Fig. 8 and Fig. 9 respectively. In the first game, the SPI player moves first. After detecting the empty board and result retrieval from the lookup table, the output illumination pattern indicates the central square as the position to put the symbol in. Then a white symbol is manually placed there. In second move, the human player places a black symbol in the left square of second row. Then two players take turns to place symbols in different empty squares. In the seventh move, the SPI player indicates placing a white symbol in the middle first row as the action. If this symbol is placed, all the symbols in the first row are white and the SPI player will become the winner. Consequently, the top square on the board showing "You lose" is also illuminated. The game terminates after the seventh move. The second game proceeds in a similar way as the first game. The difference is that both two players have not won after a maximum of nine moves are finished and all the empty squares are occupied. So the second game ends with a draw.

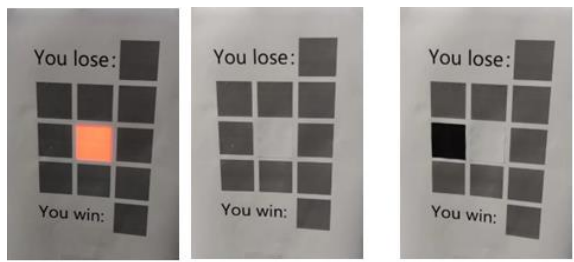

1st: SPI Player        2nd: human

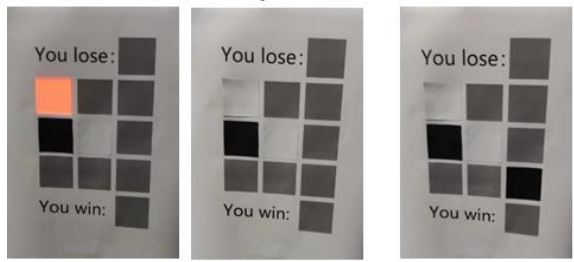

3rd: SPI Player        4th: human

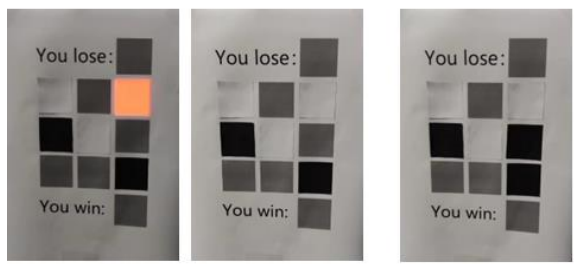

5th: SPI Player        6th: human

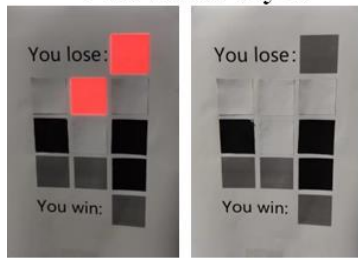

7th: SPI Player

Fig. 8 Process of first experimental game including a total of seven moves (SPI player moves first)

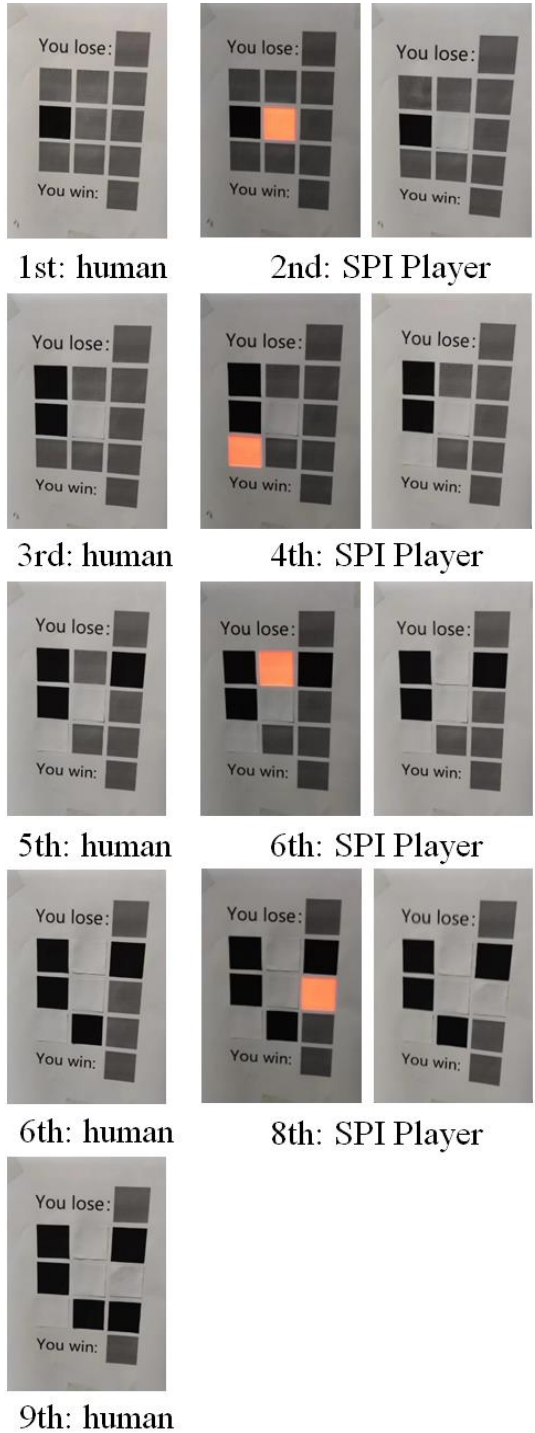

Fig. 9 Process of second experimental game including a total of nine moves (human player moves first)

In summary, an optoelectronic AI machine with minimal digital computation is implemented for playing Tic-Tac-Toe games in this work. Our proposed SPI player can detect game states by pattern illumination and single-pixel intensity recording, generate optimal actions with a lookup table and display output results by pattern illumination. The task of playing Tic-Tac-Toe games is merged into the framework of SPI and the system can defeat human players. The proposed scheme is verified by simulated and experimental results. This scheme paves the way for more diversified and interactive applications of SPI in future works.

Reference
[1] Edgar MP, Gibson GM, Padgett MJ. Principles and prospects for single-pixel imaging. Nature photonics. 2019 Jan;13(1):13-20.
[2] Duarte MF, Davenport MA, Takhar D, Laska JN, Sun T, Kelly KF, Baraniuk RG. Single-pixel imaging via compressive sampling. IEEE signal processing magazine. 2008 Mar 21;25(2):83-91.
[3] Gibson GM, Johnson SD, Padgett MJ. Single-pixel imaging 12 years on: a review. Optics Express. 2020 Sep 14;28(19):28190-208.
[4]Gong W, Zhao C, Yu H, Chen M, Xu W, Han S. Three-dimensional ghost imaging lidar via sparsity constraint. Scientific reports. 2016 May 17;6(1):1-6.
[5]Sun MJ, Zhang JM. Single-pixel imaging and its application in three-dimensional reconstruction: a brief review. Sensors. 2019 Jan;19(3):732.
[6] Lin LX, Cao J, Zhou D, Cui H, Hao Q. Ghost imaging through scattering medium by utilizing scattered light. Optics Express. 2022 Mar 28;30(7):11243-53.
[7]Le M, Wang G, Zheng H, Liu J, Zhou Y, Xu Z. Underwater computational ghost imaging. Optics express. 2017 Sep 18;25(19):22859-68.
[8] Li W, Tong Z, Xiao K, Liu Z, Gao Q, Sun J, Liu S, Han S, Wang Z. Single-frame wide-field nanoscopy based on ghost imaging via sparsity constraints. Optica. 2019 Dec 20;6(12):1515-23.
[9] Clemente P, Durán V, Tajahuerce E, Andrés P, Climent V, Lancis J. Compressive holography with a single-pixel detector. Optics letters. 2013 Jul 15;38(14):2524-7.
[10] Duan D, Zhu R, Xia Y. Color night vision ghost imaging based on a wavelet transform. Optics Letters. 2021 Sep 1;46(17):4172-5.
[11] Musarra G, Lyons A, Conca E, Altmann Y, Villa F, Zappa F, Padgett MJ, Faccio D. Non-line-of-sight three-dimensional imaging with a single-pixel camera. Physical Review Applied. 2019 Jul 18;12(1):011002.
[12] S. Jiao, J. Feng, Y. Gao, T. Lei, Z. Xie, and X. Yuan, "Optical machine learning with incoherent light and a single-pixel detector," Opt. Lett. 44(21), 5186-5189 (2019)
[13] H. Fu, L. Bian, and J. Zhang, "Single-pixel sensing with optimal binarized modulation," Opt. Lett. 45(11), 3111-3114 (2020)
[14] T. Bu, S. Kumar, H. Zhang, I. Huang, and Y.-P. Huang, "Single-pixel pattern recognition with coherent nonlinear optics," Opt. Lett. 45(24), 6771-6774 (2020)
[15] B. Limbacher, S. Schoenhuber, M. Wenclawiak, M. A. Kainz, A. M. Andrews, G. Strasser, J. Darmo, and K. Unterrainer, "Terahertz optical machine learning for object recognition," APL Photonics 5(12), 126103 (2020).
[16] Ren H, Zhao S, Gruska J. Edge detection based on single-pixel imaging. Optics express. 2018 Mar 5;26(5):5501-11.
[17] D. Shi, K. Yin, J. Huang, K. Yuan, W. Zhu, C. Xie, D. Liu, and Y. Wang, "Fast tracking of moving objects using single-pixel imaging," Opt. Commun., 440, 155-162 (2019)
[18] S. Sun, H.-K. Hu, Y.-K. Xu, Y.-G. Li, H.-Z. Lin, and W.-T. Liu, "Simultaneously Tracking and Imaging a Moving Object under Photon Crisis," Phys. Rev. Applied 17, 024050 (2022)

[19] Wu H, Wang R, Zhao G, Xiao H, Liang J, Wang D, Tian X, Cheng L, Zhang X. Deep-learning denoising computational ghost imaging. Optics and Lasers in Engineering. 2020 Nov 1;134:106183.

[20] Wang F, Wang C, Chen M, Gong W, Zhang Y, Han S, Situ G. Far-field super-resolution ghost imaging with a deep neural network constraint. Light: Science & Applications. 2022 Jan 1;11(1):1-1.

[21] Shang R, Hoffer-Hawlik K, Wang F, Situ G, Luke GP. Two-step training deep learning framework for computational imaging without physics priors. Optics Express. 2021 May 10;29(10):15239-54.

[22] Hoshi I, Shimobaba T, Kakue T, Ito T. Single-pixel imaging using a recurrent neural network combined with convolutional layers. Optics Express. 2020 Nov 9;28(23):34069-78.

[23] Zhang K, Hu J, Yang W. Deep compressed imaging via optimized pattern scanning. Photonics research. 2021 Mar 1;9(3):B57-70.

[24] Zhang Z, Li X, Zheng S, Yao M, Zheng G, Zhong J. Image-free classification of fast-moving objects using "learned" structured illumination and single-pixel detection. Optics Express. 2020 Apr 27;28(9):13269-78.

[25] Hsu FH. Behind Deep Blue: Building the computer that defeated the world chess champion. Princeton University Press; 2002.

[26] Silver D, Schrittwieser J, Simonyan K, Antonoglou I, Huang A, Guez A, Hubert T, Baker L, Lai M, Bolton A, Chen Y. Mastering the game of go without human knowledge. nature. 2017 Oct;550(7676):354-9.

[27] Beck J. Combinatorial games: Tic-Tac-Toe theory. Cambridge: Cambridge University Press; 2008 Jan 1.

[28] Elstner M, Schiller A. Playing Tic-Tac-Toe with a sugar-based molecular computer. Journal of Chemical Information and Modeling. 2015 Aug 24;55(8):1547-51.

[29] Dewdney A. A tinkertoy computer that plays Tic-Tac-Toe. Scientific American. 1989 Oct 1;261(4):120-123.

[30] Guerrin C, Aidibi Y, Sanguinet L, Leriche P, Aloise S, Orio M, Delbaere S. When light and acid play Tic-Tac-Toe with a nine-state molecular switch. Journal of the American Chemical Society. 2019 Nov 14;141(48):19151-60.

[31] D.-Z. Du, and M. P. Panos, Minimax and applications. Vol. 4. Springer Science & Business Media, 1995.